\pdfoutput=1

\documentclass[11pt]{article}

\usepackage{acl}

\usepackage{times}
\usepackage{latexsym}

\usepackage[T1]{fontenc}

\usepackage[utf8]{inputenc}

\usepackage{microtype}
\usepackage{graphicx}
\usepackage{multirow}
\usepackage{caption}
\usepackage{booktabs}

\usepackage{url}
\usepackage{xcolor}
\usepackage{epsfig}
\usepackage{adjustbox}
\usepackage{amsfonts, amsmath, amssymb}
\usepackage{booktabs} 
\usepackage{comment}
\usepackage{caption, subcaption}
\usepackage{textcomp}
\usepackage{relsize}
\usepackage{stmaryrd}
\usepackage{bbm}
\usepackage{rotating}
\usepackage{helvet}
\usepackage{courier}
\usepackage{cleveref}
\usepackage{xspace}
\usepackage{enumitem}
\usepackage{mdframed}
\usepackage{makecell}
\usepackage{tcolorbox}
\usepackage{nicematrix}
\usepackage{tabu}
\usepackage{colortbl}
\usepackage{xcolor}
\PassOptionsToPackage{table}{xcolor}

\title{Make Compound Sentences Simple to Analyze: Learning~to~Split~Sentences~for Aspect-based Sentiment Analysis}

\author{Yongsik Seo$^{\ast}$, Sungwon Song$^{\ast}$, Ryang Heo\thanks{\ \ Equal contribution}\ , Jieyong Kim, Dongha Lee\thanks{\ \ Corresponding author}\\
  Department of Artificial Intelligence, Yonsei University\\
  \texttt{\{ndata,sungwonok,ryang1119,jieyong99,donalee\}@yonsei.ac.kr}}
  
\usepackage{graphicx}
\usepackage{multirow}
\usepackage{pdfpages}

\begin{document}
\maketitle

\newcommand{\restfift}{Rest15\xspace}
\newcommand{\restsixt}{Rest16\xspace}
\newcommand{\laptopfift}{Laptop15\xspace}

\newcommand{\absa}{ABSA\xspace}
\newcommand{\asqp}{ASQP\xspace}
\newcommand{\acos}{ACOS\xspace}
\newcommand{\tasd}{TASD\xspace}
\newcommand{\aste}{ASTE\xspace}

\newcommand{\pap}{\textit{plug-and-play}\xspace}

\newcommand{\simsent}{\textit{{simple sentence}}\xspace}
\newcommand{\comsent}{\textit{{compound sentence}}\xspace}

\newcommand{\paraphrase}{Paraphrase\xspace}
\newcommand{\mvp}{MvP\xspace}
\newcommand{\ilo}{ILO\xspace}
\newcommand{\dlo}{DLO\xspace}
\newcommand{\scrap}{SCRAP\xspace}
\newcommand{\gptthree}{GPT-3.5-turbo\xspace}
\newcommand{\gptfour}{GPT-4-turbo\xspace}
\newcommand{\gptfouro}{GPT-4o\xspace}
\newcommand{\proposed}{\textsc{AToss}\xspace}

\begin{abstract}
In the domain of Aspect-Based Sentiment Analysis (ABSA), generative methods have shown promising results and achieved substantial advancements. However, despite these advancements, the tasks of extracting sentiment quadruplets, which capture the nuanced sentiment expressions within a sentence, remain significant challenges. In particular, compound sentences can potentially contain multiple quadruplets, making the extraction task increasingly difficult as sentence complexity grows. To address this issue, we are focusing on simplifying sentence structures to facilitate the easier recognition of these elements and crafting a model that integrates seamlessly with various \absa tasks.
In this paper, we propose \textbf{A}spect \textbf{T}erm \textbf{O}riented \textbf{S}entence \textbf{S}plitter (\proposed), which simplifies compound sentence into simpler and clearer forms, thereby clarifying their structure and intent. As a \pap module, this approach retains the parameters of the \absa model while making it easier to identify essential intent within input sentences. Extensive experimental results show that utilizing \proposed outperforms existing methods in both \asqp and \acos tasks, which are the primary tasks for extracting sentiment quadruplets.\footnote{Our code is available at \url{https://github.com/ryang1119/ATOSS.git}}

\end{abstract}

\section{Introduction}
\label{sec:intro}
\begin{figure}
    \centering
    \includegraphics[width=\columnwidth]{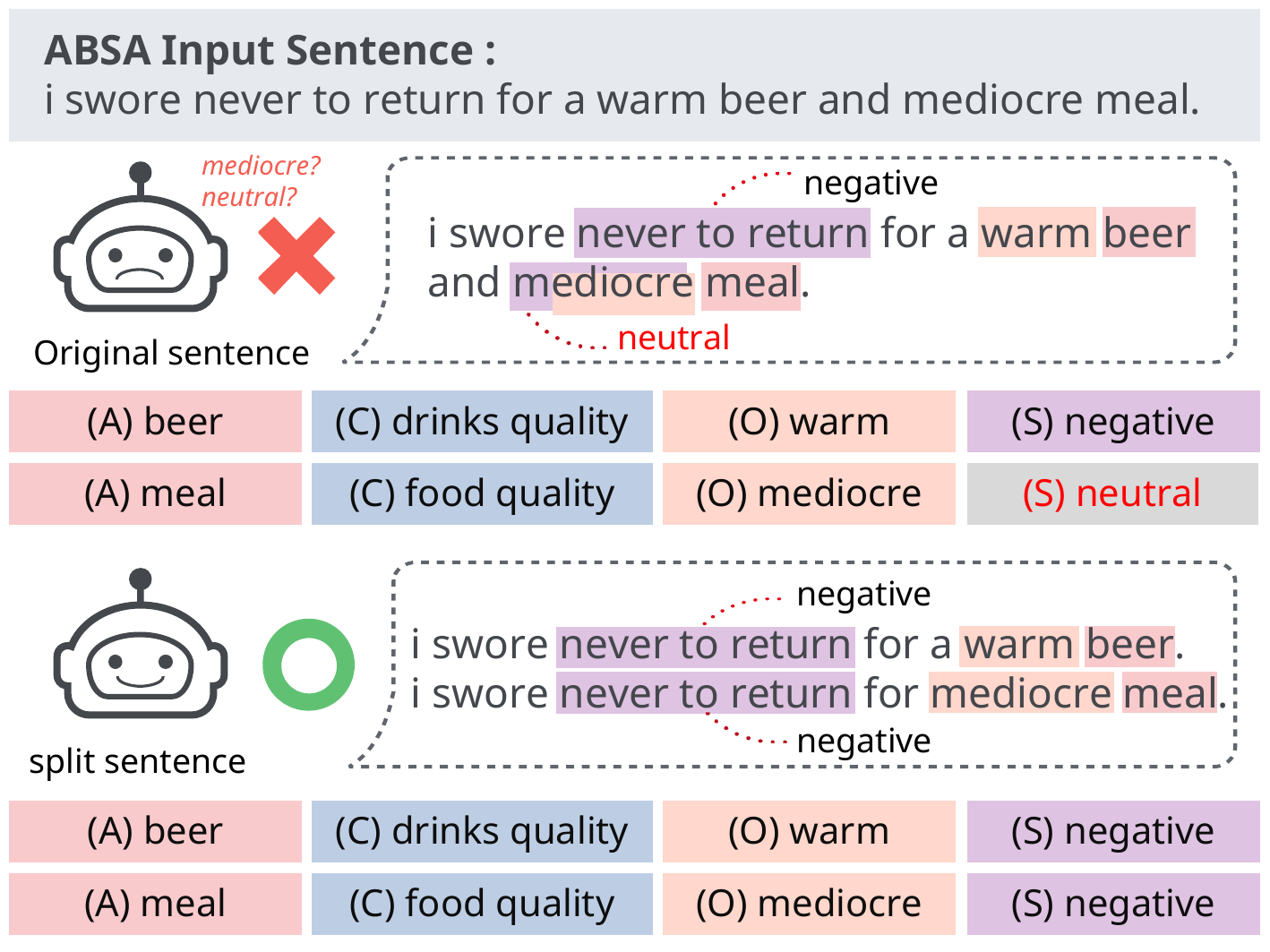} 
    \caption{Existing \absa models struggle to accurately predict quadruplets in sentence with compound syntactic structures but perform well when the sentences are provided in simpler and clearer forms.}
    \label{fig:motivation}
\end{figure}

Aspect-Based Sentiment Analysis (\absa) \cite{pontiki-etal-2014-semeval} refers to
the crucial task of understanding sentiments at the aspect-level. This technique identifies specific aspects of entities and evaluates their associated sentiments, providing richer contextual insights. In recent years, \absa has progressed beyond simple sentiment classification to tackle more complex structures like sentiment triplets and quadruplets, which include aspect term, aspect category, opinion term and sentiment polarity for quadruplets. Among these, the Aspect Sentiment Quad Prediction (\asqp) and Aspect-Category-Opinion-Sentiment (\acos) tasks, which involve predicting comprehensive sentiment quadruplets from a given sentence, are currently the most challenging tasks in \absa and are being actively researched. Recently, generative methods have been proposed as solutions for predicting quadruplets, gaining significant attention in research due to their simplicity in addressing this problem in an end-to-end manner. \cite{zhang2021aspect, hu2022improving, gou2023mvp} propose a framework that uses a sequence-to-sequence learning to transform an input sentence into predetermined output formats for predicting  quadruplets. 

Despite the state-of-the-art performances,  \absa models still suffer from the ambiguity of aspect-level sentiments in complex sentence structure, often due to multiple subjects with different states or context-dependent changes in a single subject. For example, in Figure~\ref{fig:motivation} upper, the sentence ``\textit{I swore never to return for a warm beer and mediocre meal}'' conveys a negative sentiment with ``\textit{never to return}''.
However, the opinion word ``\textit{mediocre}'' describing the ``\textit{meal}'' adds confusion to the overall sentiment of the sentence. As depicted in Figure~\ref{fig:motivation} lower, by splitting this sentence into ``\textit{I swore never to return for a warm beer}'' and ``\textit{I swore never to return for a mediocre meal}'', we can clearly identify the intended sentiment quadruplets. Additionally, sentences that involve lists, conjunctions, causal relationships, or storytelling elements are difficult to interpret clearly, as they require complex reasoning and judgment, similar to human thought processes. To address the aforementioned challenges, our research aims to enable \absa models to more easily identify intent within simple and clear sentence structures, avoiding confusion in complex and intertwined sentence constructions.

Motivated by these observations, we propose a model, named \underline{\textbf{A}}spect \underline{\textbf{T}}erm \underline{\textbf{O}}riented \underline{\textbf{S}}entence \underline{\textbf{S}}plitter (\proposed), which helps to accurately identify quadruplets by simplifying an original, compound sentence into simpler and clearer forms for \absa models. Moreover, \proposed, as a \pap module, can be integrated into each \absa model keeping their parameters. In other words, once \proposed is pre-trained, it can be immediately applied to any \absa model without the need for additional training. Specifically, \proposed is first optimized via LLM distillation, and then aligned with the target \absa model’s sentence preference (Figure~\ref{fig:overview}). We first obtain split sentences by prompting with LLM and train our model to generate the split sentence given an original sentence. Moreover, we address any ambiguities or splitting biases by further tailoring our \proposed for the target \absa model that will perform quadruplet prediction. To this end, we adopt preference alignment~\cite{rafailov2023direct} with sentence pairs of preferred-dispreferred splitting results. As a result, \proposed is fine-tuned to be further enhance the target \absa model's quadruplet prediction accuracy.

Our extensive experiments on main aspect quadruplet prediction tasks, including \asqp and \acos demonstrate that \proposed significantly enhances the prediction accuracy of state-of-the-art \absa models, including fine-tuned models and prompt-based LLMs. Specifically, \proposed effectively reduces the error rates associated with incorrectly predicting aspect terms by facilitating the identification of aspect terms in input sentences for each sentiment quadruplet. Furthermore, our \proposed splitter can be seamlessly integrated into the inference stage for other \absa tasks, highlighting its high level of generalizability. \\

Our contributions are summarized as follows:
\begin{itemize}
    \item We propose \proposed splitter which splits compound sentences into simpler and clearer forms, allowing \absa models to easily identify intent within the sentence structure.
    \item \proposed aligns with the sentence preference of target \absa model and, as a \pap module, can be seamlessly integrated into existing models to enhance performance without the need to update their parameters.
    \item Experiments show that integrating \proposed improves the quad prediction accuracy of existing \absa models while also enabling them to adapt well to other \absa tasks.
\end{itemize}

\section{Related Work}
\label{sec:relwork}
In this section, we review the existing literature on (1) state-of-the-art approaches to Aspect-Based Sentiment Analysis (ABSA), and (2) recent efforts to distillation of large language models’ (LLMs) remarkable reasoning ability on various tasks into small language models (LMs).
\subsection{Sentiment Quadruplet Prediction} 
\label{subsec:relatedwork_1}
\absa has been the focus of extensive research in recent years, aiming to extract sentiment-related elements for more fine-grained sentiment analysis. Earlier work primarily focused on predicting single or dual sentiment elements~\cite{Ma2019ExploringSL, Zhang2020ConvolutionOH}. As the field evolved, more challenging ABSA tasks were proposed, such as Aspect Sentiment Triplet Extraction (ASTE) ~\cite{peng2020knowing} and Target Aspect Sentiment Detection (TASD) ~\cite{Wan2020TargetAspectSentimentJD}, which focus on predicting sentiment triplet, and Aspect Category Opinion Sentiment (ACOS)~\cite{cai2021aspect} and Aspect Sentiment Quad Prediction (ASQP)~\cite{zhang2021aspect}, which target sentiment quadruplet.

To tackle sentiment quad prediction problems, early approaches proposed pipeline methods ~\cite{cai2021aspect}, but more recently, generative methods have emerged as the primary research focus because of their simplicity and end-to-end approach. ~\citet{zhang2021aspect} solve the \asqp task by transforming target quads into natural language sentences, using the knowledge from the pre-trained generative model, which leads to better performance. ~\citet{hu2022improving} was the first to investigate element ordering based on the quad structure, and proposed a method for predicting quads by augmenting the targets of the ASQP dataset with various permutations. Building on this, ~\citet{gou2023mvp} introduces an element order-based prompt learning method that improves sentiment tuple prediction by aggregating multi-view results.

Despite the promising results, long and complex text poses significant challenges for the model in predicting quadruplet. In this paper, we focus on a strategy for providing simpler sentences that allow \absa models to handle complexity of sentences more accurately and effectively.

\subsection{Distillation of LLM's Reasoning Ability}
\label{subsec:relatedwork_2}
Knowledge distillation, which trains smaller models based on larger models, aims to reduce their size and latency while maintaining accuracy and generalization capabilities \cite{Hinton2015DistillingTK, Sanh2019DistilBERTAD}. Large Language Models (LLMs) have demonstrated an emergent ability in reasoning by generating explanations through Chain-of-Thought (CoT) prompting \cite{wei2022chain, wang2023selfconsistency, kojima2022large}. With the remarkable performance of LLMs across a wide range of tasks, recent research has focused on distilling their reasoning capabilities into smaller language models. Fine-tune-CoT \cite{Ho2022LargeLM} has demonstrated outstanding performance across various tasks by enabling LLMs to generate diverse reasoning paths and distill them into LMs.
Some studies have conducted more detailed, task-specific knowledge distillation using LLMs \cite{Magister2022TeachingSL, Chae2023DialogueCD, Hsieh2023DistillingSO}. Recently, there has been an attempt to leverage LLM’s CoT reasoning to address imprecise predictions and limited interpretability in the ASQP task \cite{kim2024self}.

\section{Preliminaries}
\label{sec:prelim}
In this section, we formally define our target \absa tasks and analyze \absa models' behavior based on the structural complexity of input sentences.

\begin{table}[t]
    \centering
    \resizebox{0.99\linewidth}{!}{
        \begin{tabular}{ccccc}
            \toprule
            \textbf{Task} & \textbf{Datasets} & \textbf{Ratio of S / C} & \textbf{Acc of S} & \textbf{Acc of C} \\
            \midrule
            \multirow{2}{*}{\asqp} & Rest15 & 32.93 / 67.07 & 53.57 & 40.08 \\
            & Rest16 & 32.63 / 67.37 & 57.87 & 50.21 \\
            \midrule
            \multirow{2}{*}{\acos} & Laptop16 & 32.72 / 67.28 & 39.28 & 30.23  \\
            & Rest16 & 32.16 / 67.84 & 54.45 & 48.25  \\
            \bottomrule
        \end{tabular}
    }
    \caption{Proportion of \textit{simple} (S) / \textit{compound} (C) inputs, and quad prediction accuracy (i.e., Recall) of existing \absa models for \textit{simple} and \textit{compound} sentences.}
    \label{tbl:prelimanal}
\end{table}

\subsection{Problem Formulation}
\label{subsec:problem}

In this work, we focus on tasks of predicting aspect-level sentiment from input sentence, in the form of structured quadruplets, (i.e., \asqp and \acos).
Formally, given an input sentence, these task aim to predict all aspect sentiment quadruplets consisting of four components, i.e., \{(\texttt{at}, \texttt{ac}, \texttt{ot}, \texttt{sp})\}.
The aspect term ``\texttt{at}'' and opinion term ``\texttt{ot}'' are detected within the sentence, while the aspect category ``\texttt{ac}'' and sentiment polarity ``\texttt{sp}'' are classified within their respective pre-defined sets.

If the target aspect term is not explicitly mentioned, it is implicitly expressed and mapped to \textit{``NULL''}. Note that \acos differs from \asqp in its definition of the opinion term, focusing on more implicit aspects and opinions, which may result in the opinion term being \textit{``NULL''} as well.
The aspect category is classified as an element within the category set that is pre-defined for each domain or dataset;
for example, in restaurant review datasets, it includes various categories such as ``\textit{food prices}'' and ``\textit{ambience general}''.
The sentiment polarity is predicted as one of the three sentiment classes: ``\textit{positive}'', ``\textit{neutral}'', and ``\textit{negative}'', each indicating the corresponding aspect-level sentiment. 

\subsection{Analysis of ABSA Performance based on Sentence Structural Complexity}
\label{subsec:sentcomplexity}

We first investigate the prediction accuracy of existing \absa models according to the degree of complexity in an input sentence structure.
To this end, we categorize all test inputs into two sets: \textit{\simsent} and \textit{\comsent}.
In this work, we define \simsent as a sentence containing only a single independent clause and annotated with a single aspect quadruplet. In contrast, \comsent is connected by conjunctions such as ``\textit{and}'', ``\textit{or}'', ``\textit{but}'', and punctuated with commas; it can be annotated with one or more aspect quadruplets. Each sentence in the \asqp and \acos datasets may contain several quadruplets for a single aspect or multiple aspects.

\begin{figure}[t]
    \centering
    \includegraphics[width=\linewidth]{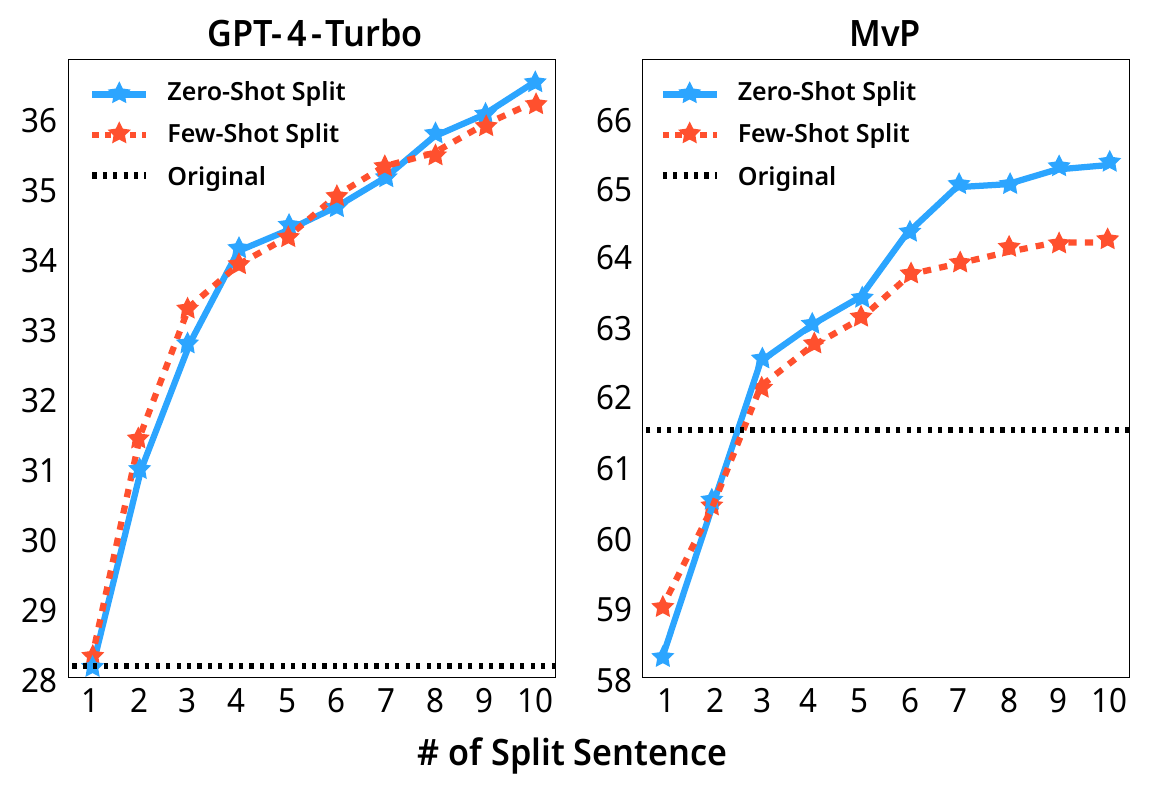}
    \caption{Performance changes of ABSA models (Left: \gptfour, Right: \mvp) w.r.t. the number of candidate split sentences. (Task: ACOS, Dataset: Rest16)}
    \label{fig:oraclevoting}
\end{figure}

Table~\ref{tbl:prelimanal} reports the error rates of state-of-the-art \absa models for both \textit{simple} and \comsent inputs. 
From the results, we observe a high error rate in \comsent across all tasks.
As illustrated in Figure~\ref{fig:motivation}, when multiple quadruplets exist for multiple aspect terms in a sentence, it becomes more complex and intricately intertwined.
In such cases, the model struggles to detect each term accurately, especially when the distance between the aspect term and sentiment component in text is relatively large, making accurate quadruplet prediction difficult.
This issue arises often when multiple aspects each have multiple associated opinions or sentiment quadruplets within the same sentence.
Based on these observations, we hypothesize that splitting a \comsent into simple ones based on aspect terms could reduce the error rate and lead to higher overall accuracy. We provide more detailed analysis in Section \ref{subsec:rqonetwo}.

In addition, we conduct a preliminary study to validate our hypothesis that simplifying sentences via splitting can enhance their clarity and thereby improve the quadruplet prediction accuracy of \absa models. To this end, we examine how much F1 score in quadruplet prediction improves when using split sentences instead of the original ones. 
Specifically, for each test input, we generate 10 candidate sentence splits by prompting LLM to perform sentence splitting in both zero-shot and few-shot manners. 
We then select the best split sentence that yields the highest F1 score (i.e., \textit{oracle} voting). Figure~\ref{fig:oraclevoting} shows the changes in prediction F1 score as the number of candidate sentence splits increases.
Both the fine-tuned model (i.e., \mvp) and the prompting-based LLM (i.e., \gptfour) demonstrate significant performance improvements with split sentences, especially when more candidates of split sentences are available.
Notably, split sentences obtained through zero-shot prompting, which result in more diverse split forms compared to few-shot prompts, show great potential for enhancing the performance of existing \absa models.
Consequently, we confirm that the strategy of splitting sentences into appropriate simple forms can indeed aid in enhancing \absa performance.

\begin{figure*}[t]
    \centering
    \includegraphics[width=\textwidth]{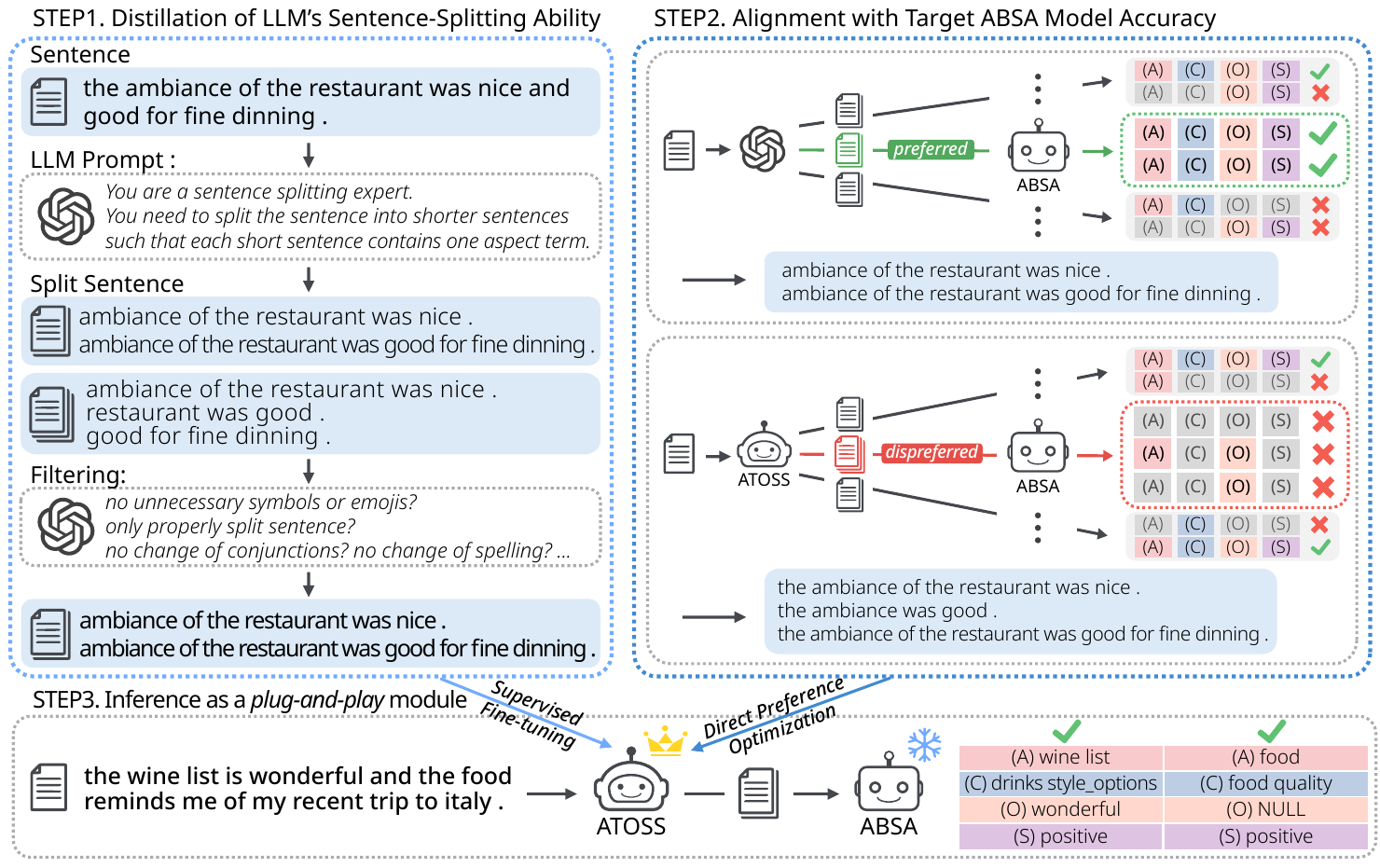}
    \caption{Overall framework for training and utilizing \proposed for \absa tasks. The training process involves (1) distillation of LLM's capability for sentence splitting, and (2) alignment with a target model's sentence preference. The inference process (3) predicts the quadruplets by taking sentences split by \proposed as the input. \proposed, as a \pap module, can enhance prediction accuracy without requiring updates to the target model’s parameters.}
    \label{fig:overview}
\end{figure*}

\section{Methodology}
\label{sec:method}
In this section, we present a \pap module that splits an input sentence into multiple simple sentences that facilitates \absa tasks. Our proposed model, named \textbf{A}spect \textbf{T}erm \textbf{O}riented \textbf{S}entence \textbf{S}plitter (\proposed), is a small LM trained via knowledge distillation from a teacher LLM and further refined to align its output with helpfulness for enhancing the target \absa model's accuracy. The overview of \proposed is illustrated in Figure~\ref{fig:overview}.

\subsection{Aspect-Oriented Splitting Strategy}
\label{subsec:stepone_}
Our research aims to enable \absa models to easily find intent within simple and clear sentence structures. To achieve this goal, we employ \textit{Aspect-Oriented Splitting Strategy}, which ensures that the split sentences contain aspect terms. As mentioned in Section ~\ref{subsec:sentcomplexity}, existing datasets consist of both \textit{simple} and \textit{compound} sentences for \absa models to process. For the \simsent, which has a straightforward structure and allows the model to accurately understand its intent without additional steps, we retain it without modification. However, the \comsent often have a complex structure, such as multiple quadruplets oriented from a single aspect term, making it challenging for existing models to discern the intent within sentence. To address this issue, we only split \comsent employing the \textit{Aspect-Oriented Splitting} strategy.

\subsection{Distillation of LLM's Splitting Ability}
\label{subsec:stepone}
We first optimize our sentence splitter to generate diverse split sentences given an original input sentence. 
To train the sentence splitter, we augment the \absa training dataset of sentence-quadruplet pairs $(s, \mathcal{Q})$ into $\mathcal{X} = \{(s, s', \mathcal{Q})\}$, where $s'$ is the split sentence from the original sentence $s$.

\paragraph{Split sentence generation}
To distill a teacher LLM's sentence splitting ability into \proposed, we generate training data for distillation by prompting the LLM.
In this step, we simply adopt zero-shot prompting for split sentence generation, allowing the LLM to diversify split $s$ into $s'$ using its pre-trained knowledge.
To explore effective and diverse $s'$, we instruct the LLM to generate 10 diverse $s'$ for each $s$ with given matching spellings.

\paragraph{Split sentence selection}
However, as $s'$ may still contain noise, we introduce an additional filtering process to select \textit{K} split sentences that best meet the specific criteria, within the set of generated $s'$.
This filtering process is also performed via LLM prompting with manually written splitting criteria.  

\paragraph{Supervised fine-tuning (SFT)}
We train the model to predict the target $s'$ given an input $s$.
With the input-target pair $(s, s')$, we fine-tune the sequence-to-sequence LM by minimizing the
following negative log-likelihood loss:
\begin{equation*}
\begin{aligned}
\tiny
\mathcal{L}_{NLL} = -\log p(s'|s) = -\sum_{t=1}^{T} \log p(s'_t|s, s'_{<t}),
\end{aligned}
\end{equation*}
where $T$ is the length of the target sequence $s$ and $s'_{<t}$ denotes previously generated tokens.
Note that the purpose of this step is to obtain a general sentence splitter by distilling an LLM's sentence splitting ability;
therefore, we use all $(s, s')$ samples collected from multiple available datasets.

\subsection{Alignment with Sentence Preference}
\label{subsec:steptwo}
Even though the general splitter is able to split the sentence to some extent, but the general split sentence may not always be an optimally processed input for a specific model.
To further refine the splitter for a target dataset, task, and \absa model, we additionally tune \proposed based on preference alignment strategy.
While shorter sentences are generally easier for tasks than longer ones, the optimal format for split sentences may vary depending on the models and datasets.
To address this issue, we apply Direct preference optimization (DPO) ~\cite{rafailov2023direct} to the \proposed for each specific model and dataset to perform the tasks.

\paragraph{Preferred sentence selection}
We obtain $s'$ that is optimal based on our manual splitting strategy (Section~\ref{subsec:sentcomplexity}), but this may exclude other optimal $s'$ formats that we do not consider.
To mitigate this issue, we utilize few-shot prompting for split sentence generation through LLM, allowing it to effectively split $s$ into $s'$ via in-context learning of given examples.
By providing LLM with $(s, \mathcal{Q})$ pair, we guide it to split $s$ based on its ground-truth aspect terms of $\mathcal{Q}$.
To explore effective and appropriate $s'$, we prompt LLM to generate 10 diverse $s'$ for each $s$, as input to measure the sentence-level F1 score from the \absa model's inference stage. We use this comparison to select the preferred sentences: if $s$ is \simsent, we select $s'$ where the number of split sentences matches the number of quadruplets; for \comsent, if $s$ has the higher F1 score, no preferred sentence is chosen; if the scores are equal, $s$ is retained; if $s$ has the lower score, we select all distinct $s'$ with higher score.

\paragraph{Dispreferred sentence selection}
Even if we construct the appropriate dataset, \proposed may fail to generate optimal $s'$ for \absa model. To alleviate this issue, we utilize beam search feature of \proposed to generate 10 different $s'$. We use this comparison to select the dispreferred sentences: if $s$ is \simsent, we select $s'$ with the hightest similarity; for \comsent, if $s$ has the lower F1 score, no dispreferred sentence is chosen; if the scores are equal, we select $s'$ with the lowest similarity to $s$; if $s$ has the higher score, we select $s'$ with the highest similarity to $s$ among those with lower score.

\paragraph{Direct preference optimization}
Through the aforementioned process, we can construct preference pairs \( P = \{(s, p^+, p^-)\} \) using the preferred sentence \( p^+ \) and the dispreferred sentence \( p^- \).
We apply DPO on our sentence splitter \( \theta \) to train a preference-tuned sentence splitter \( \theta^* \) that minimizes the following objective:
\begin{equation*}
\begin{aligned}
L_{\text{DPO}}(\theta^*; \theta) = \\
- \mathbb{E}_{(s, p^+, p^-) \sim P} & \log \sigma\left[r(s, p^+) - r(s, p^-)\right],
\end{aligned}
\end{equation*}
where \( r(s, p) = \frac{p_{\theta^*} (p|s)}{p_{\theta} (p|s)} \).
By optimizing the model using preferred-dispreferred sentence pairs, our obtained model \( \theta^* \) is trained to prefer sentences that are clearly split based on the aspect term while avoiding those that are ambiguously or unclearly split.
Note that \( \theta \) has been specifically trained for target \absa model, since the preferred split sentences vary across different models.

\subsection{Applying \proposed as a \pap}
During inference, our final \proposed model, obtained via a two-step optimization process, transforms an input sentence into a split one, which is then provided to the \absa model.
This stage adopts \proposed tailored for a specific \absa model and dataset.
By utilizing \proposed as a \pap module, existing \absa models can process input sentences optimized for each task without updating their parameters, thereby improving performance. In other words, as \absa models themselves are not required any tuning, \proposed can be universally applied to any off-the-shelf \absa models, or to any \absa tasks focusing on aspect-level sentiments. Note that it can also be adopted to closed-source LLMs, such as \gptthree, \gptfour and \gptfouro which cannot be tuned, demonstrating outstanding applicability and flexibility.

\section{Experiments}
\label{sec:exp}
We conducted our experiments while addressing the following research questions:

\begin{itemize}[leftmargin=*,topsep=2pt,itemsep=2pt,parsep=0pt]
    \item \textbf{RQ1:} Can \proposed enhance the quad prediction accuracy of existing \absa models?
    \item \textbf{RQ2:} Can \proposed improve aspect-level F1 in quad prediction for existing \absa models?
    \item \textbf{RQ3:} Can \proposed be effective for other \absa tasks beyond quad prediction?
\end{itemize}

\begin{table*}[thbp]
\centering
\resizebox{0.99\linewidth}{!}{
\begin{tabular}{lcccccccccccc}
\toprule
\multirow{4}{*}{\textbf{Methods}} & \multicolumn{6}{c}{\textbf{ASQP}} & \multicolumn{6}{c}{\textbf{ACOS}} \\
\cmidrule(lr){2-7} \cmidrule(lr){8-13}
& \multicolumn{3}{c}{\textbf{Rest15}} & \multicolumn{3}{c}{\textbf{Rest16}} & \multicolumn{3}{c}{\textbf{Laptop16}} & \multicolumn{3}{c}{\textbf{Rest16}} \\
\cmidrule(lr){2-4}\cmidrule(lr){5-7}\cmidrule(lr){8-10}\cmidrule(lr){11-13}
& \textbf{Pre} & \textbf{Rec} & \textbf{F1} & \textbf{Pre} & \textbf{Rec} & \textbf{F1} & \textbf{Pre} & \textbf{Rec} & \textbf{F1} & \textbf{Pre} & \textbf{Rec} & \textbf{F1} \\ \midrule
\multicolumn{13}{c}{\textbf{\textit{Fine-tuned models}}} \\ \midrule
\paraphrase~\cite{zhang2021aspect} & 43.70 & 47.55 & 45.54 & 56.28 & 59.45 & 57.82 & 43.23 & 42.89 & 43.06 & 58.74 & 60.55 & \underline{59.63}  \\
\quad + \proposed (\textit{General})\textsuperscript & 45.44 & 47.04 & \underline{46.23} & 57.28 & 59.57 & \underline{58.40} & 43.49 & 43.15 & \underline{43.32} & 57.84 & 59.12 & 58.47 \\
\quad + \proposed (\textit{Specific}) \textsuperscript & 46.06 & 47.80 & \textbf{46.91} & 58.05 & 60.45 & \textbf{59.23} & 43.79 & 43.76 & \textbf{43.77} & 59.01 & 62.21 & \textbf{60.57} \\ \midrule
\ilo~\cite{hu2022improving} & 48.51 & 49.06 & 48.78 & 55.98 & 60.95 & 58.36 & 45.10 & 45.56 & 45.33 & 56.32 & 57.51 & 56.91 \\
\quad + \proposed (\textit{General}) & 50.45 & 49.31 & \textbf{49.87} & 57.62 & 61.08 & \underline{59.30} & 44.51 & 44.70 & \underline{45.55} & 57.05 & 57.33 & \underline{57.33} \\
\quad + \proposed (\textit{Specific}) & 49.42 & 48.43 & \underline{48.92} & 57.96 & 61.08 & \textbf{59.48} & 46.19 & 46.43 & \textbf{46.31} & 58.20 & 58.40 & \textbf{58.30} \\ \midrule
\dlo~\cite{hu2022improving} & 46.88 & 49.18 & 48.00 & 57.28 & 61.08 & 59.12 & 43.65 & 43.84 & 43.75 & 59.30 & 59.96 & 59.62 \\
\quad + \proposed (\textit{General}) & 47.76 & 49.56 & \textbf{48.64} & 58.60 & 60.33 & \underline{59.40} & 43.92 & 43.58 & \underline{43.99} & 59.87 & 60.07 & \underline{59.97} \\
\quad + \proposed (\textit{Specific}) & 48.01 & 48.68 & \underline{48.34} & 61.50 & 58.57 & \textbf{60.15} & 45.13 & 43.93 & \textbf{44.52} & 61.18 & 59.96 & \textbf{60.56} \\ \midrule
\mvp~\cite{gou2023mvp} & 49.81 & 48.68 & 49.24 & 61.33 & 62.33 & \underline{61.82} & 43.76 & 43.69 & 43.72 & 61.27 & 57.86 & \underline{59.52} \\
\quad + \proposed (\textit{General})\textsuperscript & 51.99 & 49.18 & \underline{50.55} & 61.61 & 61.45 & 61.53 & 45.21 & 42.91 & \underline{44.03} & 61.05 & 56.99 & 58.95 \\
\quad + \proposed (\textit{Specific})\textsuperscript & 51.99 & 49.18 & \textbf{50.55} & 62.80 & 61.70 & \textbf{62.25} & 45.32 & 43.17 & \textbf{44.22} & 63.12 & 58.30 & \textbf{60.61} \\ \midrule
\multicolumn{13}{c}{\textbf{\textit{Prompting-based LLMs}}} \\ \midrule

\gptthree & 15.38 & 16.73 & 16.02 & 21.23 & 23.28 & 22.21 & 7.27 & 7.92 & 7.58 & 21.46 & 23.80 & 22.57 \\
\quad + \proposed (\textit{General}) & 20.15 & 23.14 & \textbf{21.55} & 26.13 & 29.66 & \underline{27.78} & 8.53 & 9.39 & \textbf{8.94} & 23.13 & 26.31 & \underline{24.62} \\
\quad + \proposed (\textit{Specific}) & 20.07 & 22.89 & \underline{21.39} & 27.42 & 30.79 & \textbf{29.01} & 8.33 & 9.22 & \underline{8.75} & 23.40 & 26.42 & \textbf{24.82} \\ \midrule
\gptfour & 20.11 & 25.96 & 22.66 & 23.76 & 28.16 & 25.77 & 9.24 & 10.51 & 9.83 & 28.44 & 30.68 & 28.99 \\
\quad + \proposed (\textit{General}) & 21.11 & 25.91 & \underline{23.26} & 23.52 & 28.79 & \underline{25.89} & 9.31 & 10.77 & \textbf{9.98} & 28.18 & 31.66 & \underline{29.82} \\
\quad + \proposed (\textit{Specific}) & 21.59 & 26.67 & \textbf{23.86} & 25.73 & 30.91 & \textbf{28.08} & 9.16 & 10.85 & \underline{9.94} & 28.68 & 31.88 & \textbf{30.20} \\ \midrule
\gptfouro & 18.30 & 20.88 & 19.51 & 24.05 & 27.03 & 25.46 & 10.48 & 11.28 & 10.87 & 21.63 & 21.72 & 21.68 \\
\quad + \proposed (\textit{General}) & 25.88 & 29.43 & \underline{27.55} & 32.05 & 35.54 & \underline{33.71} & 11.44 & 12.83 & \underline{12.10} & 27.63 & 28.71 & \underline{28.16} \\
\quad + \proposed (\textit{Specific}) & 26.25 & 30.31 & \textbf{28.14} & 33.22 & 36.92 & \textbf{34.97} & 11.83 & 13.35 & \textbf{12.55} & 28.14 & 29.37 & \textbf{28.74} \\ \midrule

\end{tabular}
}

\caption{Performance (\%) of various \absa models and the ones equipped with \proposed.}
\label{tbl:mainresult}
\end{table*}

\subsection{Experimental Settings}
\label{subsec:expset}
\paragraph{Tasks and datasets}
We validate the effectiveness of \proposed on 4 datasets across 2 tasks, \asqp and \acos. 
For \asqp, we utilize two restaurant domain datasets, i.e., Rest15 and Rest16 ~\cite{pontiki2015semeval, pontiki2016semeval}. 
In the case of \acos, we adopt restaurant-\acos and laptop-\acos datasets, i.e., Rest16 and Laptop16 ~\cite{cai2021aspect}. 
Refer to Appendix~\ref{subsec:Dataset statistics} for more details.

\paragraph{Implementation details}
We employ the T5-base model ~\cite{raffel2020exploring} from Huggingface Transformers\footnote{\url{https://github.com/huggingface/transformers}}~\cite{wolf2020transformers} as the backbone model of our splitter.
We adopt the \pap module in our experiments, maintaining the existing parameters of target \absa model while only tuning the parameters of the \proposed model.
To filter out noisy sentences and select those that best meet the criteria, we set \textit{K}=2.
We use two variants of our model:
the splitter trained by using only LLM Distillation, named \textit{General} (Section~\ref{subsec:stepone}), and the one aligned for preference, named  \textit{Specific} (Section~\ref{subsec:steptwo}). More details about implementation for our model are given in Appendix~\ref{subsec:Implement Details}

\paragraph{Evaluation metrics}
For all tasks, a sentiment quadruplet is considered correct if and only if every element matches gold quadruplet exactly. 
We utilize F1 score as primary evaluation metric ~\cite{zhang2021aspect_forum, mao-etal-2022-seq2path}, with all reported F1 score for fine-tuned models (i.e. \mvp) using randomly selected seeds. 
We use same F1 score metric to evaluate in a single run for prompting-based models (i.e. \gptfour).We also report precision (Pre) and recall (Rec) scores.

\paragraph{\absa models}
For \absa models, we use fine-tuned models that have recently shown outstanding performance, i.e., \textbf{\paraphrase}~\cite{zhang2021aspect}, \textbf{\ilo} \& \textbf{\dlo}~\cite{hu2022improving} and \textbf{\mvp}~\cite{gou2023mvp}. 
We also use the prompting-based LLMs, i.e., \textbf{\gptthree} (\texttt{gpt-3.5-turbo-0125}), \textbf{\gptfour} (\texttt{gpt-4-1106-preview}) and \textbf{\gptfouro} (\texttt{gpt-4o}) \footnote{\url{https://chat.openai.com/}}, employing zero-shot prompting.

\subsection{Effectiveness of \proposed (RQ1 \& RQ2)}
\label{subsec:rqonetwo}

\paragraph{Performance comparison}
\label{subsubec:rqeone}
Table~\ref{tbl:mainresult} presents \asqp and \acos performance of various models.
Overall, \absa models integrating our \proposed, which takes split sentences as inputs, lead to better performance compared to those that use original sentences as inputs. 
This improvement is consistently observed across fine-tuned models and prompting-based LLM;
this highlights efficacy of our sentence splitting approach in enhancing the accuracy of existing \absa models without parameter tuning and extensive modifications. Both the \textit{General} and \textit{Specific} versions of ATOSS yield improved results, demonstrating its consistent ability to generate clear split sentences for \absa models.

\begin{table}[t]
\centering
\resizebox{0.99\linewidth}{!}{
\begin{tabular}{lcccc}
\toprule
\multirow{2.5}{*}{\textbf{Methods}} & \multicolumn{2}{c}{\textbf{ASQP}} & \multicolumn{2}{c}{\textbf{ACOS}} \\ 
\cmidrule(lr){2-3}\cmidrule(lr){4-5}
 & \textbf{Rest15} & \textbf{Rest16} & \textbf{Laptop16} & \textbf{Rest16}  \\ \midrule
\gptthree & 16.02 & 22.21 & 7.58 & 22.57 \\
\quad + \proposed  (\textit{General}) & \textbf{21.55} & \underline{27.78} & \textbf{8.94} & \underline{24.62} \\
\quad + \proposed  (\textit{Specific})  & \underline{21.39} & \textbf{29.01} & 8.75 & \textbf{24.82} \\
\quad w / CoT Splitting & 17.84 & 23.70 & \underline{8.89} & 22.48 \\ \midrule
\gptfouro & 19.51 & 25.46 & 10.87 & 21.68 \\
\quad + \proposed  (\textit{General}) & \underline{27.55} & \underline{33.71} & \underline{12.10} & \underline{28.16} \\
\quad + \proposed  (\textit{Specific}) & \textbf{28.14} & \textbf{34.97} & \textbf{12.55} & \textbf{28.74} \\
\quad w / CoT Splitting & 21.96 & 27.08 & 12.07 & 24.87 \\ \bottomrule
\end{tabular}
}
\caption{Comparative performance (\%) of LLMs equipped with \proposed and those utilizing split-then-quadruplet prediction via zero-shot CoT prompting.}
\label{tbl:cot}
\end{table}

\begin{table*}[thbp]
\centering
\resizebox{0.99\linewidth}{!}{
\begin{tabular}{lcccccccccccc}
\toprule
\multirow{4}{*}{\textbf{Methods}} & \multicolumn{6}{c}{\textbf{ASQP}} & \multicolumn{6}{c}{\textbf{ACOS}} \\
\cmidrule(lr){2-7} \cmidrule(lr){8-13}
& \multicolumn{3}{c}{\textbf{Rest15}} & \multicolumn{3}{c}{\textbf{Rest16}} & \multicolumn{3}{c}{\textbf{Laptop16}} & \multicolumn{3}{c}{\textbf{Rest16}} \\
\cmidrule(lr){2-4}\cmidrule(lr){5-7}\cmidrule(lr){8-10}\cmidrule(lr){11-13}
& \textbf{S} & \textbf{C} & \textbf{T} & \textbf{S} & \textbf{C} & \textbf{T} & \textbf{S} & \textbf{C} & \textbf{T} & \textbf{S} & \textbf{C} & \textbf{T} \\ \midrule
\paraphrase & 86.21 & \underline{68.30} & \textbf{71.98} & 90.91 & \underline{74.89} & 78.27 & 89.39 & 73.10 & 76.80 & 91.92 & 76.01 & 79.00  \\
\quad + \proposed (\textit{Specific}) \textsuperscript & 86.44 & 67.61 & 71.55 & 91.01 & 75.25 & \textbf{78.61} & 89.38 & \underline{74.40} & \textbf{77.90} & 91.60 & \underline{76.88} & \textbf{79.79} \\ \midrule
\mvp & 80.33 & 63.24 & 67.17 & 91.88 & 73.36 & 77.44 & 92.45 & \underline{80.52} & \textbf{83.30} & 85.86 & 75.14 & 77.43\\
\quad + \proposed (\textit{Specific})\textsuperscript & 84.92 & \underline{70.31} & \textbf{73.64} & 90.20 & \underline{78.15} & \textbf{80.86} & 89.86 & 75.29 & 78.85 & 91.10 & \underline{78.99} & \textbf{81.61} \\ \midrule
\gptfouro & 42.37 & 51.48 & 49.59 & 45.51 & 57.42 & 54.92 & 69.20 & 51.67 & 55.50 & 47.83 & 60.49 & 57.95 \\
\quad + \proposed (\textit{Specific}) & 79.66 & \underline{61.61} & \textbf{65.34} & 86.52 & \underline{68.37} & \textbf{72.20} & 80.30 & \underline{63.92} & \textbf{67.42} & 84.47 & \underline{67.38} & \textbf{70.73} \\ \midrule
\end{tabular}
}
\caption{Performance (\%) of \absa models and one equipped with \proposed in terms of aspect-level F1. categorized by sentence type: \textit{simple} (S), \textit{compound} (C), and \textit{total} (T).}
\label{tbl:sim_com}
\end{table*}

\paragraph{LLM's capability in quad prediction: a comparative analysis with \proposed}
We conduct experiments that leverage zero-shot CoT prompting \cite{kojima2022large} to make the LLM analyze the structure of sentences, split them, and then predict quads from the sentences, without using our \proposed splitter. 
As shown in Table~\ref{tbl:cot}, applying \proposed significantly improves quad prediction performance, whereas relying on the LLMs' reasoning ability without \proposed results in a smaller improvement. Therefore, the effectiveness of \proposed in predicting quads has been demonstrated.



\paragraph{Aspect-level performance analysis}
\label{subsubsec:rqtwo}

We observed in Figure~\ref{fig:oraclevoting} that splitting sentences into simpler forms enhances the detection of aspect terms. Based on this observation, we further investigate whether using \proposed improves the performance of aspect term extraction in \absa models. As indicated in Table~\ref{tbl:sim_com}, we can observe an overall improvement in aspect-level F1 scores not only for the \comsent targeted by our \proposed but also for the \textit{total sentence}. In other words, these findings indicate that splitting sentences helps \absa models more effectively recognize primary intents within the input sentence structure.

\begin{table}[t]
\centering
\resizebox{0.99\linewidth}{!}{
\begin{tabular}{lcccc}
\toprule
\multirow{2.5}{*}{\textbf{Methods}} & \multicolumn{2}{c}{\textbf{TASD}} & \multicolumn{2}{c}{\textbf{ASTE}} \\ 
\cmidrule(lr){2-3}\cmidrule(lr){4-5}
 & \textbf{Rest15} & \textbf{Rest16} & \textbf{Rest15} & \textbf{Rest16}  \\ \midrule
\mvp & \underline{63.46} & \underline{71.23} & 63.29 & 73.09 \\
\quad + \proposed (\textit{General})\textsuperscript & 63.32 & 70.26 & \underline{64.66} & \underline{73.34} \\
\quad + \proposed (\textit{Specific})\textsuperscript & \textbf{64.02} & \textbf{71.33} & \textbf{65.60} & \textbf{74.06}  \\ \midrule
\gptthree & 34.66 & 40.83 & \underline{42.41} & 51.43 \\
\quad + \proposed (\textit{General})\textsuperscript & \textbf{37.28} & \textbf{41.87} & 42.14 & \underline{51.60} \\
\quad + \proposed (\textit{Specific})\textsuperscript & \underline{36.77} & \underline{41.67} & \textbf{42.90} & \textbf{53.52} \\ \midrule
\gptfouro & 50.43 & 53.80 &  49.04 & 54.82 \\
\quad + \proposed (\textit{General})\textsuperscript & \textbf{51.46} & \textbf{55.80} & \underline{49.59} & \underline{55.58} \\
\quad + \proposed (\textit{Specific})\textsuperscript & \underline{50.93} & \underline{55.79} & \textbf{50.78} & \textbf{56.99} \\
\bottomrule
\end{tabular}
}
\caption{Performance (\%) of various ABSA models equipped with \proposed in the \textit{cross-task} setting. \proposed is trained for ASQP then tested for TASD and ASTE.}
\label{tbl:crossresult}
\end{table}
\subsection{Generalizability of \proposed in \absa (RQ3)}
\label{subsec:rqthree}

We evaluate our approach in \textit{cross-task} setting by utilizing \proposed, trained for quadruplet prediction task, to preprocess the inputs of target \absa model performing the triplet prediction task, \tasd and \aste. 
When evaluating performance, we integrate a \proposed (\textit{General}) trained on ASQP with a \proposed (\textit{Specific}) tailored to the restaurant dataset from each task. 
As shown in Table~\ref{tbl:crossresult}, the results consistently demonstrate performance improvements across all datasets, highlighting the effectiveness of sentence simplification through splitting in \textit{cross-task} setting. 
These findings suggest that even when the elements to be predicted from the sentence differ across tasks, \proposed remains effective, showcasing its versatility and utility in various scenarios.

\subsection{Potential performance of \absa models in split sentences}
\begin{figure}
   \centering
   \includegraphics[width=\linewidth]{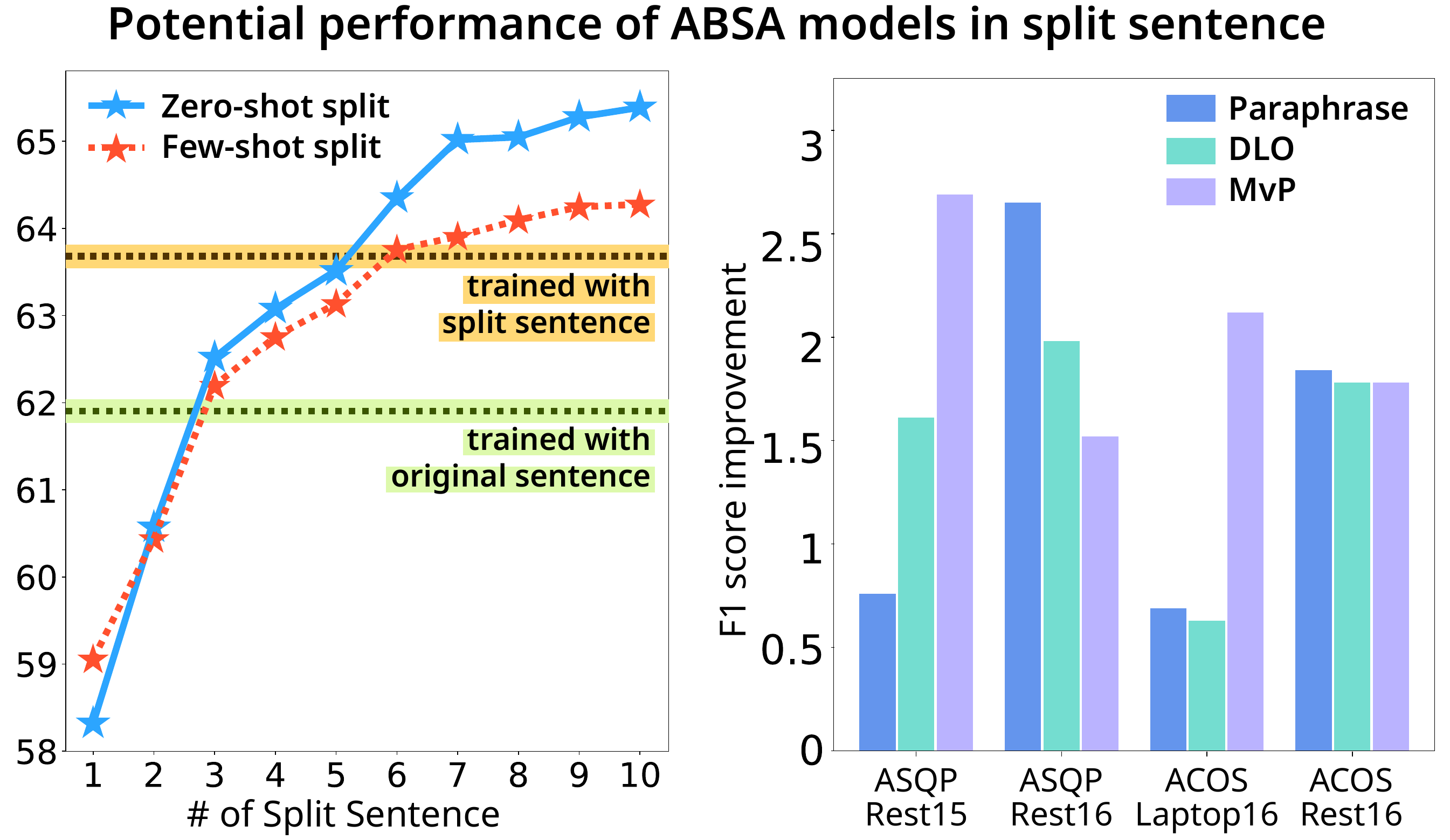}
   \caption{\textbf{Left}: Model: \mvp, Task: ACOS, Dataset: Rest16. \textbf{Right}: F1 improvement when each model is trained on split sentences instead of original sentences.}
   \label{fig:potential}
\end{figure}
To assess the potential performance of the \absa models in split sentences, we train the models using split sentences and then evaluate the inference results on split sentences. 
To this end, we collect a set of split sentences via \gptthree few-shot prompting so that it aligns well with our criteria for splitting sentence, and measure performance with this dataset. 
As illustrated in Figure~\ref{fig:potential} (\textbf{Left}), \absa models on split sentences adapts better during the inference stage to these split sentences, approaching the \textit{oracle voting} performance mentioned earlier in Figure~\ref{fig:oraclevoting}. 
Additionally, in Figure~\ref{fig:potential} (\textbf{Right}), more significant performance improvements can be achieved by using split sentences as inputs for both training and inference in models, compared to using them only for inference.
\label{subsec:potential_performance}

\section{Conclusion}
\label{sec:conclusion}
In this paper, we aim to simplify sentences through strategic splitting, allowing \absa models to better understand the inherent structure within input sentences. 
This splitting strategy, termed \textit{Aspect-Oriented Splitting}, divides sentences more concisely and clearly while retaining the essential element of the aspect term. 
Based on these points, we demonstrate that \proposed, as a \pap module, can be seamlessly integrated with various fine-tuned models as well as prompt-based LLMs, improving performance without necessitating any updates to the models themselves. 
Our research highlights the significant impact of sentence structure on sentiment analysis, presenting substantial implications for the broader field of \absa.

\section{Limitations}
\label{sec:limitations}
Despite achieving state-of-the-art performance, our study faces several limitations. 
First, the available dataset is not large enough, limiting our training effectiveness. 
The scarcity of real data for tasks such as predicting sentiment quadruplet (i.e., \asqp and \acos) required cross-training with unseen data, which might have yielded better results with a more extensive dataset. 
Secondly, the high cost associated with building our \proposed presents a significant challenge. 
We use LLMs not only to generate split sentences but also for evaluation purposes. 
Especially, we use repeated prompting for each input sentence during the data regeneration process for sentence splitting. 
Lastly, \proposed as a \pap module and is therefore dependent on existing \absa models. 
Due to this dependency, our model cannot be utilized independently and its performance may vary based on the quality and characteristics of the \absa models it uses.

\section{Ethical Statement}
\label{sec:etical}
We employ datasets that are well-recognized and previously utilized within the scientific community, ensuring both transparency and integrity in our experiments. 
In other words, our methodologies and findings do not cause harm to any individuals or groups, we have publicly released our code as open-source. 
We are aware of the potential biases in sentiment polarity predictions that may arise from using large pre-trained language models, as these models can reflect societal biases present in their training data \citep{Tan2019AssessingSA}. 
We recognize the importance of ongoing efforts to address these biases.
Moreover, we emphasize the necessity of continuous monitoring and rigorous evaluation to prevent our smaller downstream models from replicating or amplifying the biases of their larger language model counterparts.

\section*{Acknowledgements}
\label{sec:acknowledgements}
This work was supported by the IITP grants funded by the Korea government (MSIT) (No. RS-2020-II201361; RS-2024-00457882, AI Research Hub Project), and the NRF grant funded by the Korea government (MSIT) (No. RS-2023-00244689).

\bibliography{anthology,custom}

\begin{thebibliography}{27}
\expandafter\ifx\csname natexlab\endcsname\relax\def\natexlab#1{#1}\fi

\bibitem[{Cai et~al.(2021)Cai, Xia, and Yu}]{cai2021aspect}
Hongjie Cai, Rui Xia, and Jianfei Yu. 2021.
\newblock Aspect-category-opinion-sentiment quadruple extraction with implicit aspects and opinions.
\newblock In \emph{Proceedings of the 59th Annual Meeting of the Association for Computational Linguistics and the 11th International Joint Conference on Natural Language Processing (Volume 1: Long Papers)}, pages 340--350.

\bibitem[{Chae et~al.(2023)Chae, Song, iunn Ong, Kwon, Kim, Yu, Lee, Kang, and Yeo}]{Chae2023DialogueCD}
Hyungjoo Chae, Yongho Song, Kai~Tzu iunn Ong, Taeyoon Kwon, Minjin Kim, Youngjae Yu, Dongha Lee, Dongyeop Kang, and Jinyoung Yeo. 2023.
\newblock \href {https://api.semanticscholar.org/CorpusID:264146934} {Dialogue chain-of-thought distillation for commonsense-aware conversational agents}.
\newblock \emph{ArXiv}, abs/2310.09343.

\bibitem[{Gou et~al.(2023)Gou, Guo, and Yang}]{gou2023mvp}
Zhibin Gou, Qingyan Guo, and Yujiu Yang. 2023.
\newblock Mvp: Multi-view prompting improves aspect sentiment tuple prediction.
\newblock \emph{arXiv preprint arXiv:2305.12627}.

\bibitem[{Hinton et~al.(2015)Hinton, Vinyals, and Dean}]{Hinton2015DistillingTK}
Geoffrey~E. Hinton, Oriol Vinyals, and Jeffrey Dean. 2015.
\newblock \href {https://api.semanticscholar.org/CorpusID:7200347} {Distilling the knowledge in a neural network}.
\newblock \emph{ArXiv}, abs/1503.02531.

\bibitem[{Ho et~al.(2022)Ho, Schmid, and Yun}]{Ho2022LargeLM}
Namgyu Ho, Laura Schmid, and Se-Young Yun. 2022.
\newblock \href {https://api.semanticscholar.org/CorpusID:254877399} {Large language models are reasoning teachers}.
\newblock In \emph{Annual Meeting of the Association for Computational Linguistics}.

\bibitem[{Hsieh et~al.(2023)Hsieh, Li, Yeh, Nakhost, Fujii, Ratner, Krishna, Lee, and Pfister}]{Hsieh2023DistillingSO}
Cheng-Yu Hsieh, Chun-Liang Li, Chih-Kuan Yeh, Hootan Nakhost, Yasuhisa Fujii, Alexander~J. Ratner, Ranjay Krishna, Chen-Yu Lee, and Tomas Pfister. 2023.
\newblock \href {https://api.semanticscholar.org/CorpusID:258461606} {Distilling step-by-step! outperforming larger language models with less training data and smaller model sizes}.
\newblock \emph{ArXiv}, abs/2305.02301.

\bibitem[{Hu et~al.(2022)Hu, Wu, Gao, Bai, and Zhao}]{hu2022improving}
Mengting Hu, Yike Wu, Hang Gao, Yinhao Bai, and Shiwan Zhao. 2022.
\newblock Improving aspect sentiment quad prediction via template-order data augmentation.
\newblock \emph{arXiv preprint arXiv:2210.10291}.

\bibitem[{Kim et~al.(2024)Kim, Heo, Seo, Kang, Yeo, and Lee}]{kim2024self}
Jieyong Kim, Ryang Heo, Yongsik Seo, SeongKu Kang, Jinyoung Yeo, and Dongha Lee. 2024.
\newblock Self-consistent reasoning-based aspect-sentiment quad prediction with extract-then-assign strategy.
\newblock \emph{arXiv preprint arXiv:2403.00354}.

\bibitem[{Kojima et~al.(2022)Kojima, Gu, Reid, Matsuo, and Iwasawa}]{kojima2022large}
Takeshi Kojima, Shixiang~Shane Gu, Machel Reid, Yutaka Matsuo, and Yusuke Iwasawa. 2022.
\newblock Large language models are zero-shot reasoners.
\newblock \emph{Advances in neural information processing systems}, 35:22199--22213.

\bibitem[{Ma et~al.(2019)Ma, Li, Wu, Xie, and Wang}]{Ma2019ExploringSL}
Dehong Ma, Sujian Li, Fangzhao Wu, Xing Xie, and Houfeng Wang. 2019.
\newblock \href {https://api.semanticscholar.org/CorpusID:196177821} {Exploring sequence-to-sequence learning in aspect term extraction}.
\newblock In \emph{Annual Meeting of the Association for Computational Linguistics}.

\bibitem[{Magister et~al.(2022)Magister, Mallinson, Adamek, Malmi, and Severyn}]{Magister2022TeachingSL}
Lucie~Charlotte Magister, Jonathan Mallinson, Jakub Adamek, Eric Malmi, and Aliaksei Severyn. 2022.
\newblock \href {https://api.semanticscholar.org/CorpusID:254823156} {Teaching small language models to reason}.
\newblock \emph{ArXiv}, abs/2212.08410.

\bibitem[{Mao et~al.(2022)Mao, Shen, Yang, Zhu, and Cai}]{mao-etal-2022-seq2path}
Yue Mao, Yi~Shen, Jingchao Yang, Xiaoying Zhu, and Longjun Cai. 2022.
\newblock \href {https://doi.org/10.18653/v1/2022.findings-acl.174} {{S}eq2{P}ath: Generating sentiment tuples as paths of a tree}.
\newblock In \emph{Findings of the Association for Computational Linguistics: ACL 2022}, pages 2215--2225, Dublin, Ireland. Association for Computational Linguistics.

\bibitem[{Peng et~al.(2020)Peng, Xu, Bing, Huang, Lu, and Si}]{peng2020knowing}
Haiyun Peng, Lu~Xu, Lidong Bing, Fei Huang, Wei Lu, and Luo Si. 2020.
\newblock Knowing what, how and why: A near complete solution for aspect-based sentiment analysis.
\newblock In \emph{Proceedings of the AAAI conference on artificial intelligence}, volume~34, pages 8600--8607.

\bibitem[{Pontiki et~al.(2015)Pontiki, Galanis, Papageorgiou, Manandhar, and Androutsopoulos}]{pontiki2015semeval}
Maria Pontiki, Dimitrios Galanis, Harris Papageorgiou, Suresh Manandhar, and Ion Androutsopoulos. 2015.
\newblock Semeval-2015 task 12: Aspect based sentiment analysis.
\newblock In \emph{Proceedings of the 9th international workshop on semantic evaluation (SemEval 2015)}, pages 486--495.

\bibitem[{Pontiki et~al.(2016)Pontiki, Galanis, Papageorgiou, Androutsopoulos, Manandhar, AL-Smadi, Al-Ayyoub, Zhao, Qin, De~Clercq et~al.}]{pontiki2016semeval}
Maria Pontiki, Dimitris Galanis, Haris Papageorgiou, Ion Androutsopoulos, Suresh Manandhar, Mohammed AL-Smadi, Mahmoud Al-Ayyoub, Yanyan Zhao, Bing Qin, Orph{\'e}e De~Clercq, et~al. 2016.
\newblock Semeval-2016 task 5: Aspect based sentiment analysis.
\newblock In \emph{ProWorkshop on Semantic Evaluation (SemEval-2016)}, pages 19--30. Association for Computational Linguistics.

\bibitem[{Pontiki et~al.(2014)Pontiki, Galanis, Pavlopoulos, Papageorgiou, Androutsopoulos, and Manandhar}]{pontiki-etal-2014-semeval}
Maria Pontiki, Dimitris Galanis, John Pavlopoulos, Harris Papageorgiou, Ion Androutsopoulos, and Suresh Manandhar. 2014.
\newblock \href {https://doi.org/10.3115/v1/S14-2004} {{S}em{E}val-2014 task 4: Aspect based sentiment analysis}.
\newblock In \emph{Proceedings of the 8th International Workshop on Semantic Evaluation ({S}em{E}val 2014)}, pages 27--35, Dublin, Ireland. Association for Computational Linguistics.

\bibitem[{Rafailov et~al.(2023)Rafailov, Sharma, Mitchell, Ermon, Manning, and Finn}]{rafailov2023direct}
Rafael Rafailov, Archit Sharma, Eric Mitchell, Stefano Ermon, Christopher~D Manning, and Chelsea Finn. 2023.
\newblock Direct preference optimization: Your language model is secretly a reward model.
\newblock \emph{arXiv preprint arXiv:2305.18290}.

\bibitem[{Raffel et~al.(2020)Raffel, Shazeer, Roberts, Lee, Narang, Matena, Zhou, Li, and Liu}]{raffel2020exploring}
Colin Raffel, Noam Shazeer, Adam Roberts, Katherine Lee, Sharan Narang, Michael Matena, Yanqi Zhou, Wei Li, and Peter~J Liu. 2020.
\newblock Exploring the limits of transfer learning with a unified text-to-text transformer.
\newblock \emph{The Journal of Machine Learning Research}, 21(1):5485--5551.

\bibitem[{Sanh et~al.(2019)Sanh, Debut, Chaumond, and Wolf}]{Sanh2019DistilBERTAD}
Victor Sanh, Lysandre Debut, Julien Chaumond, and Thomas Wolf. 2019.
\newblock \href {https://api.semanticscholar.org/CorpusID:203626972} {Distilbert, a distilled version of bert: smaller, faster, cheaper and lighter}.
\newblock \emph{ArXiv}, abs/1910.01108.

\bibitem[{Tan and Celis(2019)}]{Tan2019AssessingSA}
Yi~Chern Tan and Elisa Celis. 2019.
\newblock \href {https://api.semanticscholar.org/CorpusID:202781363} {Assessing social and intersectional biases in contextualized word representations}.
\newblock \emph{ArXiv}, abs/1911.01485.

\bibitem[{Wan et~al.(2020)Wan, Yang, Du, Liu, Qi, and Pan}]{Wan2020TargetAspectSentimentJD}
Hai Wan, Yufei Yang, Jianfeng Du, Yanan Liu, Kunxun Qi, and Jeff~Z. Pan. 2020.
\newblock \href {https://api.semanticscholar.org/CorpusID:214354571} {Target-aspect-sentiment joint detection for aspect-based sentiment analysis}.
\newblock In \emph{AAAI Conference on Artificial Intelligence}.

\bibitem[{Wang et~al.(2023)Wang, Wei, Schuurmans, Le, Chi, Narang, Chowdhery, and Zhou}]{wang2023selfconsistency}
Xuezhi Wang, Jason Wei, Dale Schuurmans, Quoc Le, Ed~Chi, Sharan Narang, Aakanksha Chowdhery, and Denny Zhou. 2023.
\newblock \href {http://arxiv.org/abs/2203.11171} {Self-consistency improves chain of thought reasoning in language models}.

\bibitem[{Wei et~al.(2022)Wei, Wang, Schuurmans, Bosma, Xia, Chi, Le, Zhou et~al.}]{wei2022chain}
Jason Wei, Xuezhi Wang, Dale Schuurmans, Maarten Bosma, Fei Xia, Ed~Chi, Quoc~V Le, Denny Zhou, et~al. 2022.
\newblock Chain-of-thought prompting elicits reasoning in large language models.
\newblock \emph{Advances in Neural Information Processing Systems}, 35:24824--24837.

\bibitem[{Wolf et~al.(2020)Wolf, Debut, Sanh, Chaumond, Delangue, Moi, Cistac, Rault, Louf, Funtowicz et~al.}]{wolf2020transformers}
Thomas Wolf, Lysandre Debut, Victor Sanh, Julien Chaumond, Clement Delangue, Anthony Moi, Pierric Cistac, Tim Rault, R{\'e}mi Louf, Morgan Funtowicz, et~al. 2020.
\newblock Transformers: State-of-the-art natural language processing.
\newblock In \emph{Proceedings of the 2020 conference on empirical methods in natural language processing: system demonstrations}, pages 38--45.

\bibitem[{Zhang and Qian(2020)}]{Zhang2020ConvolutionOH}
Mi~Zhang and Tieyun Qian. 2020.
\newblock \href {https://api.semanticscholar.org/CorpusID:226262228} {Convolution over hierarchical syntactic and lexical graphs for aspect level sentiment analysis}.
\newblock In \emph{Conference on Empirical Methods in Natural Language Processing}.

\bibitem[{Zhang et~al.(2021{\natexlab{a}})Zhang, Deng, Li, Bing, and Lam}]{zhang2021aspect_forum}
Wenxuan Zhang, Yang Deng, Xin Li, Lidong Bing, and Wai Lam. 2021{\natexlab{a}}.
\newblock Aspect-based sentiment analysis in question answering forums.
\newblock In \emph{Findings of the Association for Computational Linguistics: EMNLP 2021}, pages 4582--4591.

\bibitem[{Zhang et~al.(2021{\natexlab{b}})Zhang, Deng, Li, Yuan, Bing, and Lam}]{zhang2021aspect}
Wenxuan Zhang, Yang Deng, Xin Li, Yifei Yuan, Lidong Bing, and Wai Lam. 2021{\natexlab{b}}.
\newblock Aspect sentiment quad prediction as paraphrase generation.
\newblock \emph{arXiv preprint arXiv:2110.00796}.

\end{thebibliography}
\bibliographystyle{acl_natbib}



\newpage

\appendix

\appendix
\section{Experiment Details}
\subsection{Software and Hardware}
\label{subsec:Software and Hardward}
We use Pytorch to implement all the models (Python 3.8). Our all experiments are conducted with a single NVIDIA A6000 with 48GB of RAM.

\subsection{Implementation Details}
\label{subsec:Implement Details}

\proposed \textit{(General)} is trained by performing Supervised fine-tuning (SFT) on  split sentences by LLM zero-shot prompting. The training batch size is set to 64, and the validation batch size is set to 8. Training is conducted for 50 epochs with the learning rate of 6e-5. Early stopping is implemented with a patience of 20 epochs. \proposed \textit{(Specific)} is trained on \proposed \textit{(General)} applying Direct preference optimization (DPO) ~\cite{rafailov2023direct} to reflect each ABSA model's preferred split sentences. Both the training and validation batch sizes are set to 8. Training is conducted for 1 epoch with the learning rate is set to 1e-4. The beta parameter is set at 0.1 and the loss function used is sigmoid.


\section{Prompts for Sentence Splitting}
\label{subsec:zero shot}
We use \gptfour (\texttt{gpt-4-1106-preview}) to generate split sentences for \proposed. Table~\ref{tab:zero shot prompt} shows the zero-shot prompt for \proposed \textit{(General)}, and Table~\ref{tab:few shot prompt} shows the few-shot prompt for \proposed \textit{(Specific)}.

\begin{table*}
    \small
    \centering
    \begin{tabular}{p{15cm}}
    \toprule
    \textbf{Prompt: \proposed (\textit{General})} \\
    \midrule
\textcolor{teal}{\textbf{[Task Description]}}\\
    You are a sentence splitting expert. You will be provided with a review sentence and a few [aspect, category, sentiment, opinion] quadruplets from that review sentence. Here is the definition of each element in the quadruplet: \\
    - The ‘aspect’ refers to a specific feature, attribute, or aspect of a product or service that a user may express an opinion about. The aspect term might be ‘null’ for an implicit aspect. \\
    - The ‘opinion’ refers to the sentiment or attitude expressed by a user towards a particular aspect or feature of a product or service. The opinion term might be ‘null’ for an implicit opinion. \\
    - The ‘category’ refers to the category that the aspect belongs to (e.g. food quality, restaurant general, etc.). \\
    - The ‘sentiment’ refers to the sentiment class of the aspect (e.g. positive, negative, neutral). 
    \\\\
    You need to split the sentence into shorter sentences such that each short sentence contains one aspect term. When splitting, sentences connected by conjunctions must be divided into individual sentences along with their conjunctions. This process must specify the subject in every sentence. This process must retain the existing spellings exactly as in the original sentence. This process must also retain the existing spacings exactly as in the original sentence. If the sentence is too short to split or does not need to be split, use the original sentence as is. No numbering, line breaks, or explanations are needed.
    \\\\
\bottomrule
    \end{tabular}
    \caption{The zero-shot prompt for the distillation of LLM's splitting ability on ABSA.}
    \label{tab:zero shot prompt}
\end{table*}

\begin{table*}
    \small
    \centering
    \begin{tabular}{p{15cm}}
    \toprule
    \textbf{Prompt: \proposed (\textit{Specific})} \\
    \midrule
\textcolor{teal}{\textbf{[Task Description]}}\\
    You are a sentence splitting expert. You will be provided with a review sentence and a few [aspect, category, sentiment, opinion] quadruplets from that review sentence. Here is the definition of each element in the quadruplet: \\
    - The ‘aspect’ refers to a specific feature, attribute, or aspect of a product or service that a user may express an opinion about. The aspect term might be ‘null’ for an implicit aspect. \\
    - The ‘opinion’ refers to the sentiment or attitude expressed by a user towards a particular aspect or feature of a product or service. The opinion term might be ‘null’ for an implicit opinion. \\
    - The ‘category’ refers to the category that the aspect belongs to (e.g. food quality, restaurant general, etc.). \\
    - The ‘sentiment’ refers to the sentiment class of the aspect (e.g. positive, negative, neutral). 
    \\\\
    You need to split the sentence into shorter sentences such that each short sentence contains one aspect term. When splitting, sentences connected by conjunctions must be divided into individual sentences along with their conjunctions. This process must specify the subject in every sentence. This process must retain the existing spellings exactly as in the original sentence. This process must also retain the existing spacings exactly as in the original sentence. If the sentence is too short to split or does not need to be split, use the original sentence as is. No numbering, line breaks, or explanations are needed.
    \\\\

    \textcolor{teal}{\textbf{[Example 1]}} \\
\textbf{Original sentence:} i will be going back and heartily recommend it !
\\\\
\textbf{Quadruplets:} [['null', 'restaurant general', 'positive', 'recommend']]
\\\\
\textbf{Split sentence:} i will be going back and heartily recommend it !
\\\\
    \textcolor{teal}{\textbf{[Example 2]}} \\
\textbf{Original sentence:} i ' ve never had bad service and the fish is fresh and delicious .
\\\\
\textbf{Quadruplets:} [['service', 'service general', 'positive', 'never had bad'], ['fish', 'food quality', 'positive', 'fresh'], ['fish', 'food quality', 'positive', 'delicious']]
\\\\
\textbf{Split sentence:} i ' ve  never had bad service . and the fish is fresh and delicious .
\\\\
    \textcolor{teal}{\textbf{[Example 3]}} \\
\textbf{Original sentence:} very immature bartender , didnt know how to make specific drinks , service was so slowwwww , the food was not fresh or warm , waitresses were busy flirting with men at the bar and werent very attentive to all the customers .
\\\\
\textbf{Quadruplets:} [['bartender', 'service general', 'negative', 'immature'], ['service', 'service general', 'negative', 'slowwwww'], ['food', 'food quality', 'negative', 'not fresh or warm'], ['waitresses', 'service general', 'negative', 'werent very attentive']]
\\\\
\textbf{Split sentence:} very immature bartender, didnt know how to make specific drinks. service was so slowwwww. the food was not fresh or warm. waitresses were busy flirting with men at the bar and werent very attentive to all the customers .
\\\\

    \textcolor{teal}{\textbf{[Example 4]}} ... \\

\bottomrule
    \end{tabular}
    \caption{The few-shot prompt for aligning with sentence preference of each ABSA model (Examples of 4 to 10 are omitted in this table).}
    \label{tab:few shot prompt}
\end{table*}

\section{Dataset Statistics}
\label{subsec:Dataset statistics}

\begin{table*}[thbp]
    \centering
\begin{tabular}{ccccc}
\toprule
\textbf{Task} &
  \textbf{Dataset (\#C)} &
  \textbf{\begin{tabular}[c]{@{}c@{}}Train\\ (POS/NEU/NEG)\end{tabular}} &
  \textbf{\begin{tabular}[c]{@{}c@{}}Dev\\ (POS/NEU/NEG)\end{tabular}} &
  \textbf{\begin{tabular}[c]{@{}c@{}}Test\\ (POS/NEU/NEG)\end{tabular}} \\ \midrule
\multirow{3}{*}{ASQP} &
  Rest15      (\#13)&
  \begin{tabular}[c]{@{}c@{}}834\\ 1,005 / 34 / 315\end{tabular} &
  \begin{tabular}[c]{@{}c@{}}209\\ 252 / 14 / 81\end{tabular} &
  \begin{tabular}[c]{@{}c@{}}537\\ 453 / 37 / 305\end{tabular} \\ \cmidrule{2-5} 
 &
  Rest16      (\#13)&
  \begin{tabular}[c]{@{}c@{}}1,264\\ 1,369 / 62 / 558\end{tabular} &
  \begin{tabular}[c]{@{}c@{}}316\\ 341 / 23 / 143\end{tabular} &
  \begin{tabular}[c]{@{}c@{}}544\\ 584 / 40 / 177\end{tabular} \\ \midrule
\multirow{3}{*}{ACOS} &
  Rest16      (\#13)&
  \begin{tabular}[c]{@{}c@{}}1,530\\ 1,656 / 95 / 733\end{tabular} &
  \begin{tabular}[c]{@{}c@{}}171\\ 180 / 12 / 69\end{tabular} &
  \begin{tabular}[c]{@{}c@{}}583\\ 668 / 44 / 205\end{tabular} \\ \cmidrule{2-5} 
 &
  Laptop16  (\#121)&
  \begin{tabular}[c]{@{}c@{}}2,934\\ 2,583 / 227 / 1,364\end{tabular} &
  \begin{tabular}[c]{@{}c@{}}326\\ 279 / 24 / 137\end{tabular} &
  \begin{tabular}[c]{@{}c@{}}816\\ 716 / 65 / 380\end{tabular} \\ \midrule

\multirow{3}{*}{ASTE} & Rest14 (-) &
  \begin{tabular}[c]{@{}c@{}}1,266\\ 1,692 / 166 / 480\end{tabular} &
  \begin{tabular}[c]{@{}c@{}}310\\ 404 / 54 / 119\end{tabular} &
  \begin{tabular}[c]{@{}c@{}}492\\ 773 / 66 / 155\end{tabular} \\ \cmidrule{2-5}
  
 & Laptop14 (-) &      
  \begin{tabular}[c]{@{}c@{}}906\\ 817 / 126 / 517\end{tabular} &
  \begin{tabular}[c]{@{}c@{}}219\\ 169 / 36 / 141\end{tabular} &
  \begin{tabular}[c]{@{}c@{}}328\\ 364 / 63 / 116\end{tabular} \\ \midrule
  
\multirow{3}{*}{TASD} &
  Rest15      (\#13)&
  \begin{tabular}[c]{@{}c@{}}1,120\\ 1,198 / 53 / 403\end{tabular} &
  \begin{tabular}[c]{@{}c@{}}10\\ 6 / 0/ 7\end{tabular} &
  \begin{tabular}[c]{@{}c@{}}582\\ 454 / 45 / 346\end{tabular} \\ \cmidrule{2-5} 
 &
  Rest16      (\#13)&
  \begin{tabular}[c]{@{}c@{}}1,708\\ 1,657 / 101 / 749\end{tabular} &
  \begin{tabular}[c]{@{}c@{}}29\\ 23 / 1 / 297\end{tabular} &
  \begin{tabular}[c]{@{}c@{}}587\\ 611 / 44 / 204\end{tabular} \\ \bottomrule

\end{tabular}
    \caption{Dataset statistics for various tasks. \#C denote the the number of aspect categories in the pre-defined set. POS, NEU, and NEG refer to the number of positive, neutral, and negative quadruplets or triplets respectively.}
    \label{tab:Dataset statistics for various tasks}
\end{table*}
Tables~\ref{tab:Dataset statistics for various tasks} presents the dataset statistics for all \asqp, \acos, \aste, and \tasd task covered in our work.

\clearpage


\end{document}